%% file: Main.tex
\documentclass[sigconf]{acmart}

\usepackage{algorithm}
\usepackage{algorithmic}

\usepackage{booktabs}
\usepackage{bm}
\usepackage{makecell}
\usepackage{multirow}
\usepackage{amsmath}
\usepackage{mathrsfs}
\usepackage{graphicx}
\usepackage{natbib}
\usepackage{color}

\AtBeginDocument{%
  \providecommand\BibTeX{{%
    \normalfont B\kern-0.5em{\scshape i\kern-0.25em b}\kern-0.8em\TeX}}}

\setcopyright{acmcopyright}
\copyrightyear{2022}
\acmYear{2022}
\acmDOI{10.1145/3534678.3539274}
\acmConference[KDD '22]{Proceedings of the 28th ACM SIGKDD Conference on Knowledge Discovery and Data Mining}{August 14--18, 2022}{Washington, DC, USA}
\acmBooktitle{Proceedings of the 28th ACM SIGKDD Conference on Knowledge Discovery and Data Mining (KDD '22), August 14--18, 2022, Washington, DC, USA}
\acmPrice{15.00}
\acmISBN{978-1-4503-9385-0/22/08}

\begin{document}

\title{Learning the Evolutionary and Multi-scale Graph Structure for Multivariate Time Series Forecasting}






\author{Junchen Ye}
\authornote{Both authors contributed equally to this work.}
\author{Zihan Liu}
\authornotemark[1]
\email{yjchen@buaa.edu.cn, liuzihan@buaa.edu.cn}
\affiliation{%
  \institution{SKLSDE Lab, Beihang University}
  \city{Beijing}
  \country{China}
  \postcode{100191}
}


\author{Bowen Du}
\email{dobowen@buaa.edu.cn}
\affiliation{
  \institution{SKLSDE Lab,\\  Beihang University}
  \city{Beijing}
  \country{China}
  \postcode{100191}
}
\additionalaffiliation{
  \institution{Peng Cheng Lab}
  \city{Shenzhen}
  \country{China}
  \postcode{518055}
}

\author{Leilei Sun}
\authornotemark[2]
\authornote{Corresponding Author}
\email{leileisun@buaa.edu.cn}
\affiliation{
  \institution{SKLSDE Lab, \\ Beihang University}
  \city{Beijing}
  \country{China}
  \postcode{100191}
}

\author{Weimiao Li}
\email{lwm568@buaa.edu.cn}
\affiliation{
  \institution{SKLSDE Lab, \\ Beihang University}
  \city{Beijing}
  \country{China}
  \postcode{100191}
}

\author{Yanjie Fu}
\email{yanjie.fu@ucf.edu}
\affiliation{
  \institution{Department of Computer Science, University of Central Florida}
  \city{FL}
  \country{USA}
  \postcode{32816}
}

\author{Hui Xiong}
\email{xionghui@ust.hk}
\affiliation{
  \institution{Hong Kong University of Science and Technology}
  \city{Hong Kong}
  \country{China}
  \postcode{999077}
}

\renewcommand{\shortauthors}{Junchen Ye et al.}
\begin{abstract}
Recent studies have shown great promise in applying graph neural networks for multivariate time series forecasting, where the interactions of time series are described as a graph structure and the variables are represented as the graph nodes. Along this line, existing methods usually assume that the graph structure (or the adjacency matrix), which determines the aggregation manner of graph neural network, is fixed either by definition or self-learning.  However, the interactions of variables can be dynamic and evolutionary in real-world scenarios. Furthermore, the interactions of time series are quite different if they are observed at different time scales. To equip the graph neural network with a flexible and practical graph structure, in this paper, we investigate how to model the evolutionary and multi-scale interactions of time series. In particular, we first provide a hierarchical graph structure cooperated with the dilated convolution to capture the scale-specific correlations among time series. Then, a series of adjacency matrices are constructed under a recurrent manner to represent the evolving correlations at each layer. Moreover, a unified neural network is provided to integrate the components above to get the final prediction. In this way, we can capture the pair-wise correlations and temporal dependency simultaneously. Finally, experiments on both single-step and multi-step forecasting tasks demonstrate the superiority of our method over the state-of-the-art approaches.
\end{abstract}


\settopmatter{printacmref=true, printfolios=false}
\maketitle


\input{1-Introduction}

\input{2-Preliminary}

\input{3-Methodology}

\input{4-Experiment}

\input{5-RelatedWork}

\input{6-Conclusion}

\input{Acknowledgements}

\bibliographystyle{ACM-Reference-Format}
\bibliography{main}

\input{7-Appendix}

\end{document}

%% file: 1-Introduction.tex
\section{Introduction}

Time series forecasting is a ubiquitous problem in practical scenarios. By modeling the evolution of the states or events in the future, it enables decision-making and plays a vital role in numerous domains, such as traffic \cite{li2017diffusion}, healthcare \cite{jin2018treatment}, and finance \cite{zhang2017stock}. The tremendous value of this problem is also proved by the long research history. In the early time, Auto-Regressive (AR) model and its variants are the most popular methods in classical statistical domain due to the efficiency and perfect mathematical properties.
However, they are mainly applied in univariate forecasting problem and assume the linear dependency among variables. 
With the rapid growth of data volume, it is difficult for AR to deal with more complicated conditions due to the relatively low expressiveness.

Multivariate time series forecasting explores the correlation among variables. Recent years have witnessed that a number of deep learning methods are applied in this domain to handle non-linear dependency. LSTNet \cite{lai2018modeling} and TPA-LSTM \cite{shih2019temporal} are the first two works toward multivariate time series forecasting based on the deep learning framework. To be more specific, they both combine the convolution neural network (CNN) and recurrent neural network (RNN) to capture the intra- and inter-time-series dependencies respectively. However, it is difficult for the global aggregation of CNN to pair-wise correlations among variables precisely. To solve this problem, graph neural network (GNN), the generalization of convolutional neural network to non-Euclidean space, comes to the stage. By treating the variables as nodes, the connections among them could be represented by edges properly. \citet{li2017diffusion} first combined the GNN with the gated recurrent unit to give the prediction and introduced a hand-craft adjacency matrix to describe the correlations based on the distance between nodes.
\citet{wu2019wavenet} argued that the predefined graph structure could not reflect the genuine connections, and they constructed a self-learned adjacency matrix during the training process. 
\citet{shang2021discrete} simplified the bilevel program problem and sampled the discrete graph structure from the Bernoulli distribution.
How to design the proper graph structure to model the correlations among the time series gradually becomes the key to solving this problem.

\begin{figure}[t]
	\centering
	\includegraphics[width=\columnwidth]{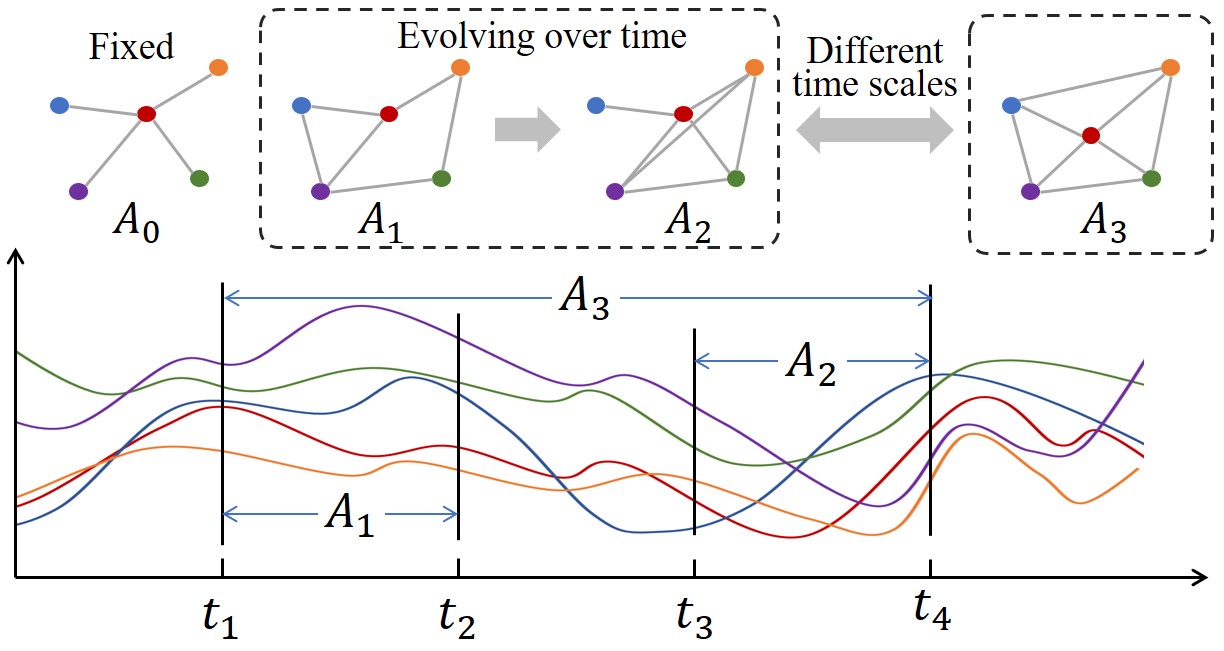}
	\caption{The possible interactions of variables in multivariate time series forecasting. Most existing works utilize the fixed correlation ($A_0$). However, the graph structure is evolving ($A_1$ and $A_2$) and varies in different observation scales ($A_3$).}
	\label{fig:motivation}
\end{figure}

Though remarkable success has been achieved by generalizing GNN to multivariate time series forecasting domain, there are still several important problems that remain to be further addressed: 
1) \textit{The graph structure is evolving over time.} As it is shown in the Figure \ref{fig:motivation}, the purple line and the blue line fluctuate together from $t_1$ to $t_2$, but they separate and head the different direction from $t_3$ to $t_4$. While most existing works maintain a fixed and static adjacency matrix from the beginning to the end, which could not handle such complicated condition obviously.
2) \textit{The graph structure varies on different observation scales.} The correlation between variables in short-term view could differ from it in long-term. 
For example, in the finance domain, two stocks might go up and down together under the influence of a new policy released by the government in a short term view, but they will part ways definitely in the long term if one company behind is thriving and the other is about to go bankrupt. The correlations between time series at different scales are seldom considered, and it is also obvious that a fixed adjacency matrix could not deal with it. Therefore, we argue that existing works have not explored and unleashed the full potential of the graph neural network for this problem.

When we propose to take a further step and address the two problems above, three challenges are faced:
1) The evolving graph structure is not only influenced by the current input but also strongly correlated to itself at the previous time step. The recurrent construction manner has been rarely discussed.
2) Generating the graph structure for each time step to model the evolution through existing self-learned methods would bring too many parameters, which results in difficulty for model convergence.
3) It is a nontrivial endeavor to capture the scale-specific graph structure among nodes due to the excess information and messy relationship behind it.

To cope with above challenges, we propose a novel deep learning framework named \textit{\textbf{E}volving Multi-\textbf{S}cale \textbf{G}raph Neural Network} (ESG). 
Specifically, a hierarchical architecture is proposed to capture the scale-specific inter-and intra-time-series correlations simultaneously cooperated with the dilated convolution module. Next, instead of maintaining a fixed graph structure all the time, for each scale, we constructed a series of adjacency matrices to model the evolving correlations with gated recurrent unit.
Last but not least, the final prediction is made by a unified forecasting model which fuses the multi-scale representations. 
The main contributions are summarized as follows:
\begin{itemize}
    \item This paper studies how to improve the GNN-based multivariate time series forecasting methods by constructing multiple evolutionary graph structures to model the interactions of time series. Most of the existing methods are founded on a fixed graph structure, which are not sufficient to capture the evolutionary interactions of time series, and not able to observe the interactions with different time scales either.
    \item Correspondingly, a temporal convolution module and an evolving structure learner are particularly designed to learn the multi-scale representations of time series and a series of recurrent graph structures respectively.
    \item Experiments on real-world data sets not only validate the effectiveness of the proposed method, but also illustrate how the interactions of time series evolve over time, and how to model the interactions of time series with multiple observational scales.

\end{itemize}

\begin{figure*}[tbh!]
	\centering
	\includegraphics[width=2\columnwidth]{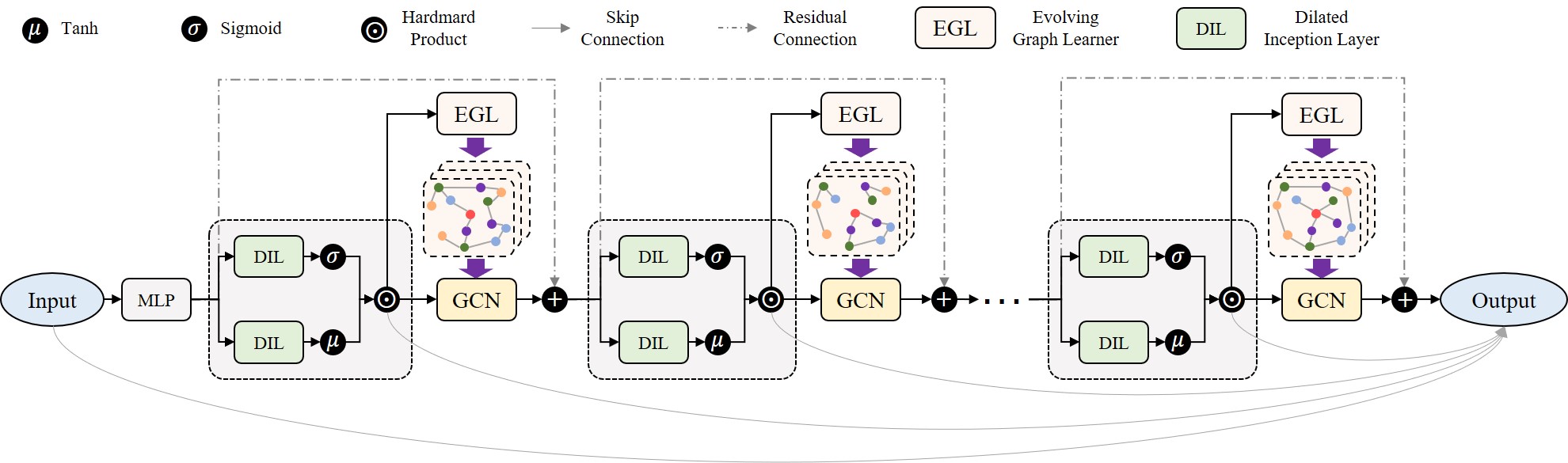}
	\caption{The framework of ESG.}
	\label{fig:model}
\end{figure*}

It is worth mentioning that the end intention of the paper is to improve the accuracy of time series forecasting, rather than discovering the ground-truth graph structure and inferring the causality. Either the hand-crafted or self-learned graph structure might contain several causal information, but they serve more likely as the external factors which help to extract global and precise signals for time series. 
We also argue that there doesn't exist a perfect and standard measurement for learned graph structure but the forecasting accuracy. For example, the genuine topological structure of the road network doesn't exploit the full potentialities of the traffic prediction problem. Additionally, the learned graph structure provides a healthy and robust version for a more accurate prediction if the "ground truth" graph exists \cite{shang2021discrete}.


%% file: 2-Preliminary.tex
\section{Preliminary}
\label{sec:pre}


This section gives a detailed definition of the multivariate time series forecasting problem.

\begin{definition}[Graph Neural Network]
We denote the relationship among all variables via a graph $G = (V, E)$, where $V$ and $E$ indicate the set of nodes and edges respectively. 
For an edge $\varepsilon \in E$, it could be represented by an ordered tuple $(v_i, v_j)$ which indicates the edge points from node $v_i$ to node $v_j$. $\mathcal{N}_i$ indicates the neighbors of node $v_i$, and all nodes connected with $v_i$ are included. The connectivity among the whole graph is represented by the adjacency matrix $\bm{A} \in \mathbb{R}^{N \times N}$ with $\bm{A}_{ij} \neq 0$ iff $(v_i,v_j) \in E$ and $\bm{A}_{ij}=0$ iff $(v_i,v_j) \notin E$, where $N$ is the total number of nodes. In order to capture the correlation between variables, the graph theory is generalized to the multivariate time series analysis domain \cite{wu2020connecting}. With denoting the variables as the nodes in graph, it is much more efficient and effective to model the correlations between variables via the adjacency matrix $\bm{A}$.
\end{definition}

\begin{definition}[Problem Formalization]
The time series with $N$ variables are denoted as $\mathbf{X} = \{ \bm{X}^{(1)}, \bm{X}^{(2)},  \cdots,  \bm{X}^{(T)} \}$ with $\mathbf{X} \in \mathbb{R}^{N \times T \times C}$, and $\bm{X}^{(t)}$ indicates the values of the variables at time step $t$. $C$ is the feature dimension of a single variable. The forecasting problem takes the historical observations to predict the states of variables in the future. According to the number of output steps, the problem setting usually falls into two mainstream, single-step, and multi-step forecasting. Given a long historical time series and a look-back window with the fixed-length $P$, the single-step forecasting proposes to obtain the future value $\bm{X}^{(t+Q)}$ at time step $Q$. In the multi-step forecasting, the historical information with fixed length $P$ is also taken into consideration, but the predicting target turns to a sequence of future values $\bm{X}^{(t+1:t+Q)}$:
\begin{equation}
\begin{aligned}
& \quad  \bm{X}^{(t-P+1:t)} \xrightarrow{\mathcal{F}_1} \bm{X}^{(t+Q)},& \\
&\bm{X}^{(t-P+1:t)} \xrightarrow{\mathcal{F}_2} \bm{X}^{(t+1:t+Q)},&
\end{aligned}
\end{equation}
where the $\mathcal{F}_1$ and $\mathcal{F}_2$ denote the mapping function that we intend to parameterize for the single-step forecasting and multi-step forecasting respectively.
\end{definition}

%% file: 3-Methodology.tex
\section{Methodology}
\label{sec:met}

In this section, the proposed framework and all components will be stated elaborately.

\subsection{Overview}
Here we introduce the overview architecture of Evolving Multi-Scare Graph Neural Network which is shown in Figure \ref{fig:model}. The multivariate time series $\bm{X}$ is first fed into a fully connected layer to obtain the initial representation $\bm{Z}^{(1)}$,
and the stacked multi-scale extractors follow. 
Each extractor is made up of three components. A temporal convolution module $\mathcal{F}_t$ is utilized to capture the multi-scale representations on the temporal dimension. The output of this part $\bm{\xi}$ is fed into the evolving graph structure learner $\mathcal{F}_a$ and the graph convolution module $\mathcal{F}_g$, which is defined as:
\begin{equation}
\begin{aligned}
\bm{\xi}^{(l)} &= \mathcal{F}_t^{(l)}(\bm{Z}^{(l)}), \\
\bm{\mathbf{A}}^{(l)} &= \mathcal{F}_a^{(l)}(\bm{\xi}^{(l)}), \\
{\bm{Z}^{\prime}}^{(l+1)} &= \mathcal{F}_g^{(l)}(\bm{\xi}^{(l)}, \bm{\mathbf{A}}^{(l)}), \\
\end{aligned}
\end{equation}
where $\bm{Z}^{(l)}$ is the input of $l$-th layer. Residual connection is employed to deliver the initial input to the next layer directly. Therefore, $\bm{Z}^{(l+1)}$ is obtained by adding ${\bm{Z}^{\prime}}^{(l+1)}$ and $\bm{Z}^{(l)}$ up. It is worth mentioning that the three modules all vary from layer to layer, which helps to extract multi-scale information.
The output of the evolving graph structure learner is a series of adjacency matrices $\bm{\mathbf{A}}^{(l)}$ which are fed into the graph convolution module $\mathcal{F}_g$.
Skip connection is utilized to deliver the information to the final output:
\begin{equation}
  \hat{\bm{Y}} \ = \mathcal{F}_o(\bm{X}, \bm{\xi}^{(1)},\bm{\xi}^{(2)},...,\bm{\xi}^{(L)}, \bm{Z}^{(L+1)}),
\end{equation}
where $L$ is the total number of the multi-scale extractor layers, and $\mathcal{F}_o$ is a simple predictor which could be implemented by a fully connected layer. Including the graph structure, ESG is trained in an end-to-end manner.
In the rest of this section, the evolving graph structure learner, temporal convolution module, and graph convolution module will be elaborated respectively. 

\begin{figure}[tbh!]
	\centering
	\includegraphics[width=\columnwidth]{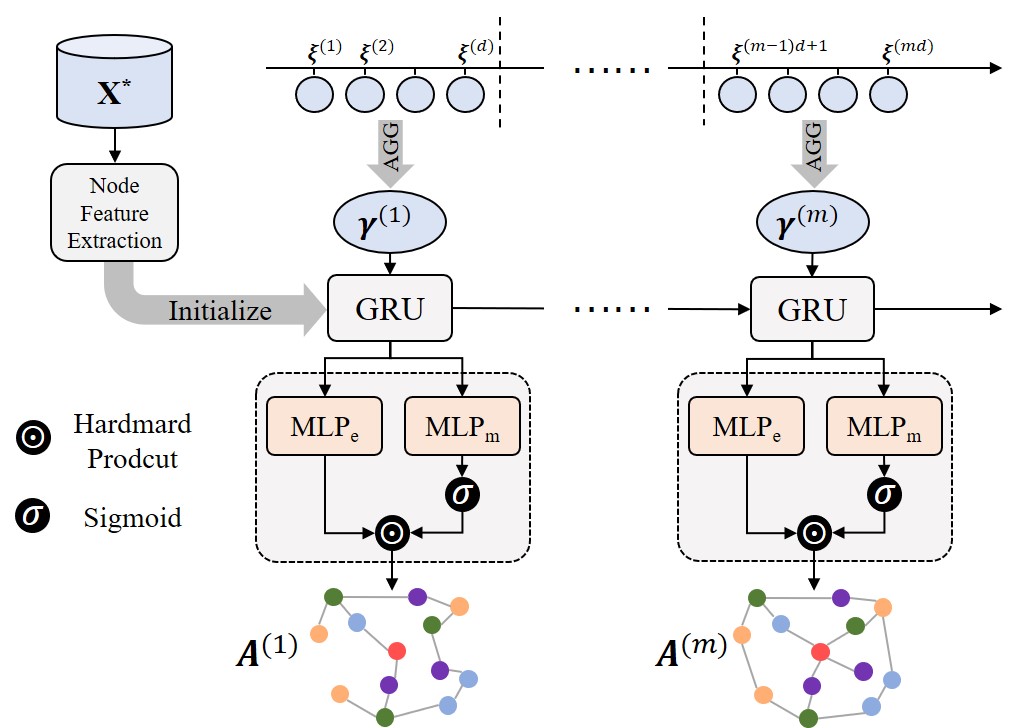}
	\caption{The detailed architecture of evolving graph structure learner (EGL).}
	\label{fig:graphstructure}
\end{figure}

\subsection{Learning Evolving Graph Structure}
\label{sec:EGL}

The correlations among multivariate time series do not stay unchanged all the time in the practical scenario. However, the dynamic correlations are seldom considered due to the complex dependency and high computational cost.
We design an \textit{\textbf{E}volving \textbf{G}raph Structure \textbf{L}earner} (EGL) to extract the dynamic correlations among variables to address this issue. The detail architecture is shown in Figure \ref{fig:graphstructure}. This module both considers the dependency with the current input values and the graph structure at last time step, which could be formulated under a recurrent manner:
\begin{equation}
\label{eq:evovle}
    \bm{A}^{(t)} = \mathcal{F}_e(\bm{A}^{(t-1)}, \bm{\xi}^{(t)}),
\end{equation}
where the $\bm{A}^{(t)} \in \mathbb{R}^{N \times N}$ denotes the adjacency matrix which describes the evolving correlations at time step $t$, $\bm{\xi}^{(t)}$ denotes the node features, and $\mathcal{F}_e$ is the evolving correlations extracting function. It is worth mentioning that we leave out the superscript $l$ which indicates the layer number in this equation and the following of this subsection for simplicity.
However, in the practical scenarios, the graph structure varies smoothly over time rather than drastically. Most of the time, adjacent timestamps follow the time consistency and have similar or even identical estimations of the relationship. Thus, we assume that the graph structures in our model remain unchanged in a time interval while having evolutionary relationships between adjacent time intervals.
Additionally, directly parameterizing $N \times N$ adjacency matrix and the mapping function $\mathcal{F}_{e}$ bring a great deal of computation cost. To address this issue, we denote that the nodes possess an evolving representation $\bm{\alpha}$ which also varies over time. And the evolving graph structure could be derived from the evolving node representations.

Here we utilize GRU, a simple but powerful variant of recurrent neural network, to model the dynamic of evolving representations.
Node features $\bm{\xi} \in \mathbb{R}^{T \times N \times C_{\bm{\xi}}}$ are divided into several segments along the temporal dimension. An aggregator is applied to the features of each segment to obtain an input sequence of GRU $[\bm{\gamma}^{(1)}, \bm{\gamma}^{(2)},...,\bm{\gamma}^{(m)},..., \bm{\gamma}^{(M)}]$:
\begin{equation}
\bm{\gamma}^{(m)} = AGG(\bm{\xi}^{((m-1)d+1 : md)}) \in \mathbb{R}^{N \times C_{\bm{\xi}}},
\end{equation}
where $d$ and $M$ are the time interval and the total number of segments. $AGG$ indicates the aggregator which could be implemented by the mean operation. Denoting $\bm{\alpha}^{(m)} \in \mathbb{R}^{N \times C_e}$ as the hidden state, the updating processing of GRU is defined as:
\begin{equation}
\begin{aligned}
\bm{r}^{(m)} & =  \sigma(\bm{W}_{r}[\bm{\gamma}^{(m)},\bm{\alpha}^{(m-1)}]+\bm{b}_{r}), \\
\bm{u}^{(m)} & =  \sigma(\bm{W}_{u}[\bm{\gamma}^{(m)},\bm{\alpha}^{(m-1)}]+\bm{b}_{u}), \\
\bm{o}^{(m)} & =  {\mu}(\bm{W}_{o}[\bm{\gamma}^{(m)},(\bm{r}^{(m)}\odot\bm{\alpha}^{(m-1)})] + \bm{b}_{o}), \\
\bm{\alpha}^{(m)} & = \bm{u}^{(m)} \odot \bm{\alpha}^{(m-1)} + (1-\bm{u}^{(m)}) \odot \bm{o}^{(m)},
\end{aligned}
\end{equation}
where $\bm{r}^{(m)}$ and $\bm{u}^{(m)}$ denote the reset gate and update gate, $\odot$ is the Hardmard product, and $\bm{W}_{r}, \bm{W}_{u}, \bm{W}_{o}$ are the learning parameters. $\sigma$ is the sigmoid function, and $\mu$ is the tangent hyperbolic function.

In the practical scenario, some intrinsic features of time series help to forecast a lot. We propose to integrate those useful static node representations $\bm{\alpha}_s$ with a fully connected layer as the initial hidden state of GRU:
\begin{equation}
\bm{\alpha}^{(0)} = {\rm MLP}_s(\bm{\alpha}_{s}).  
\end{equation}
The instability of the training process brought by the cold start problem is also handled.
However, sometimes it is not convenient to obtain the external factors. Then
we turn to the rich information which the multivariate time series themselves possess. A node feature extractor is adopted to extract static representations $\bm{\alpha}_s \in \mathbb{R}^{N \times C_s}$ from the whole training set $\bm{X^*}$ without external knowledge:
\begin{equation}
    \bm{\alpha}_{s,i} = \mathcal{F}_s(\bm{X}^*_i),
\end{equation}
where $\bm{\alpha}_{s,i}$ and $\bm{X}^*_i$ indicate the static representation and the training set data for node $i$ respectively, and $C_s$ is the number of static feature dimensions. The node feature extractor $\mathcal{F}_s$ could be implemented as many deep network structures, such as multilayer perceptron, recurrent neural network, and will be optimized during the end-to-end training process for the whole model. Additionally, the inputs to the node feature extractor are not limited to the time series data in the training set. In case the external knowledge about the attributes of nodes is given, we could introduce them to form a more comprehensive static node representation.

After the evolving node representations are generated, we concatenate the two nodes representations and apply a multilayer perceptron to derive the graph structure. Additionally, we learn a mask to control the output information ratio:
\begin{equation}
\begin{aligned}
\Hat{\bm{A}}^{(m)}_{ij} &
= {\rm MLP}_e(\bm{\alpha}^{(m)}_{i} , \bm{\alpha}^{(m)}_{j}), \\
\bm{M}^{(m)}_{ij} & 
= {\rm MLP}_m(\bm{\alpha}^{(m)}_{i} , \bm{\alpha}^{(m)}_{j}), \\
\bm{A}^{(m)} & 
= \Hat{\bm{A}}^{(m)} \odot  \sigma(\bm{M}^{(m)}),
\end{aligned}
\end{equation}
where $\Hat{\bm{A}}^{(m)}_{ij}$, $\bm{M}^{(m)}_{ij}$ denote the values of learned graph structure and the mask at the row $i$ column $j$, $\sigma$ denotes the sigmoid function. $\bm{A}^{(m)}$ is the final evolving graph structure at the $m$-th time interval.

\subsection{Temporal Convolution Module}
The temporal convolution module consists of two dilated inception layers to extract the multi-scale representations \cite{wu2020connecting}. 
Time series could possess an extreme long-term dependency in the practical scenario. We introduce the dilation factor which controls the skipping distance to the standard causal convolution. Therefore, the receptive field could expand exponentially with the increase of the layer depth. 
For the node $i$, the dilated convolution is defined as:
\begin{equation}
\bm{Z}_i \star \bm{f}_{1 \times k}(t) = \sum_{\tau = 0}^{k-1}\bm{f}_{1 \times k}(\tau)\bm{Z}_i(t-s \times \tau),
\end{equation}
where $s$ is the dilation factor, $\bm{Z}_i$ indicates the input sequence for a specific layer of node $i$, and the superscript $l$ which represents the number of layer is left out in this subsection for simplicity. $\bm{f}_{1 \times k}$ is the 1D convolution filter kernel with the size $k$. 

However, the challenge still remains that it is hard to capture both the short-term and long-term patterns simultaneously by a single filter. That the dependencies are entangled with others leads to the hard situation to discover the valuable signals. To address this problem, multiple filters with different sizes are adopted to extract temporal patterns with various ranges. Thus, the dilated inception layer is defined as:
\begin{equation}
    \bm{\xi}_i^{\prime} = concat(\bm{Z}_i \star \bm{f}_{1 \times k_1}, \bm{Z}_i \star \bm{f}_{1 \times k_2},...,\bm{Z}_i \star \bm{f}_{1 \times k_{\omega}}),
\end{equation}
where $[k_1, k_2, ..., k_{\omega}]$ are $\omega$ different filter sizes, and the outputs of different filters are truncated to the same length according to the largest filter and concatenated across the channel dimension. The gating mechanism is also utilized to control the amount of information passing to the next module. Specifically, we feed the outputs of two dilated inception layers $\bm{\xi}_{1,i}^{\prime}, \bm{\xi}_{2,i}^{\prime}$ through two different activation functions, and then make element-wise multiplication:
\begin{equation}
    \bm{\xi}_i = \sigma(\bm{\xi}_{1,i}^{\prime}) \odot \mu(\bm{\xi}_{2,i}^{\prime}),
\end{equation}
where $\sigma$ denotes the sigmoid function,  $\mu$ denotes the tangent hyperbolic function and $\odot$ is the Hadamard product. $\bm{\xi}_i$ is the output of the temporal convolution module of node $i$, and it will be fed into the evolving graph structure learner and the graph convolution module. By stacking multiple layers, the temporal convolution module captures temporal patterns at different temporal levels. For example, at the bottom layer, the module extracts short-term information while at the top layer the module tackles long-term information.

\subsection{Evolving Graph Convolution Module}

Applying the graph neural network in the multivariate time series forecasting domain has achieved great success. However, the dependency among variables not only evolves over time but also varies on different time scales, which is difficult for the fixed adjacency matrix to describe such correlations. In addition, the evolving patterns of the graph structure are also not the same at different time scales. To tackle the above problems, we utilize the scale-specific evolving graph structure learner to discover correlations among variables for the specific scale level. Formally, the output of the $l$-th temporal convolution layer $\bm{\xi}^{(l)}$ is fed into the $l$-th EGL proposed in Section \ref{sec:EGL} to generate a series of adjacency matrices as follows:
\begin{equation}
\label{eq:ip}
     [\bm{A}^{(l,1)}, \bm{A}^{(l,2)}, ...,\bm{A}^{(l,M^{(l)})}] = \mathcal{F}_a^{(l)}(\bm{\xi}^{(l)}, d^{(l)}),
\end{equation}
where $\bm{\mathbf{A}}^{(l)} = [\bm{A}^{(l,1)}, \bm{A}^{(l,2)}, ...,\bm{A}^{(l,M^{(l)})}]$, $d^{(l)}$ is the time interval, and $M^{(l)}$ determines the number of adjacency matrices at $l$-th layer.

The graph convolution $\mathcal{F}_g$ at each scale is implemented by the mix-hop propagation which consists of two steps, the information propagation, and the information selection. The former one is defined as:
\begin{equation}
\label{eq:is}
\bm{H}_{(\psi)} = \beta \bm{\xi} + (1-\beta) \bm{A}\bm{H}_{(\psi-1)},
\end{equation}
where $\xi$ indicates the input of graph convolution layer, $\bm{H}_{(\psi)}$ is the representation at hop $\psi$ and we set $\bm{H}_{(0)} = \bm{\xi}$.
$\beta$ is the hyper-parameter which controls the ratio between the original input and the information from different hops. The multi-level representations are attached with different weights adaptively:
\begin{equation}
\bm{Z}^{\prime} = \sum^{\Psi}_{\psi = 0} \bm{H}_{(\psi)} \bm{W}_{(\psi)},
\end{equation}
where $\Psi$ indicates the depth of propagation, and we also leave out the superscripts $l$ and $m$ for simplicity in Equation \eqref{eq:ip} and \eqref{eq:is}.
The information propagation step propagates node information along with the given graph structure recursively, and retain a proportion of the node's original states during the propagation process so that the propagated node states can both preserve locality and explore the deeper neighborhood, which also relieves the problem of over-smoothing to a certain extent. 
$\bm{W}_{(\psi)}$ is introduced as the feature selector to lay more importance to the hop which contains the crucial signals. 

Finally, at $l$-th layer, the representation in the $m$-th time interval will be fed into mix-hop propagation layer with its corresponding adjacency matrix $\bm{A}^{(l,m)}$, which is defined as:
\begin{equation}
    {\bm{Z}^{\prime}}^{(l,m)} = \mathcal{F}_g^{(l)}(\bm{\xi}^{(l,(m-1)d^{(l)}+1:md^{(l)})},  \bm{A}^{(l,m)}),
\end{equation}
where $\mathcal{F}_g^{(l)}$ indicates the mix-hop propagation at layer $l$. ${\bm{Z}^{\prime}}^{(l,m)}$ is the output of graph convolution at $m$-th segment, $l$-th layer, and the ${\bm{Z}^{\prime}}^{(l)}$ is the output set of all segments at $l$-th layer,  
${\bm{Z}^{\prime}}^{(l)} =  [{\bm{Z}^{\prime}}^{(l,1)}, {\bm{Z}^{\prime}}^{(l,2)} ,..., {\bm{Z}^{\prime}}^{(l,M^{(l)})}]$.

%% file: 4-Experiment.tex
\section{Experiments}
\label{sec:exp}
In this section, we verify the superiority of our model through extensive and rigorous experiments.

\subsection{Datasets \& Setup}
We conduct detailed experiments on six popular real-world datasets\footnote{Codes and datasets are available at \url{https://github.com/LiuZH-19/ESG}}. Brief statistical information is listed in Table \ref{tab:datasets}. We utilize two groups of evaluation metrics for the different forecasting tasks. For the single-step prediction, Root Relative Squared Error (RSE) and Empirical Correlation Coefficient (CORR) are selected \cite{wu2020connecting}. The multi-step prediction tasks are evaluated by Root Mean Squared Error (RMSE), Mean Absolute Error (MAE), Empirical Correlation Coefficient (CORR) \cite{ye2021coupled}. The lower value indicates better performance for all evaluation metrics except CORR. 
More datasets and setup details are stated in Appendix \ref{sec:data} and \ref{sec:setup}.

\begin{table}[h]
    \centering
     \caption{The overall information for datasets.}
    \resizebox{\columnwidth}{!}{
    \begin{tabular}{cccccc}
    \hline
    Datasets  &Nodes &Timesteps &Granularity &Task Types &Partition \\
    \hline
    
    Solar-Energy &137 &52560  &10min &\multirow{4}{*}{Single-step} &\multirow{4}{*}{6/2/2}\\
    Electricity &321 &26304 &1hour & &\\
    Exchange Rate &8 &7588 &1day & &\\
    Wind &28 &10957 &1day & &\\
    \hline
    NYC-Bike &250 &4368 &30min &\multirow{2}{*}{Multi-step} &\multirow{2}{*}{7/1.5/1.5} \\
    NYC-Taxi &266 &4368 &30min & &\\
    
    \hline
    \end{tabular}
    }
	\label{tab:datasets}
\end{table}

\begin{table*}[tbh!]           
	\centering
	\caption{Comparison with baselines on single-step forecasting.}
    \resizebox{\textwidth}{!}{
	\begin{tabular}{cc||cccc|cccc|cccc|cccc}
		\hline
		\multirow{2}{*}{Dataset} & \multirow{2}{*}{Metrics} & \multicolumn{4}{c}{Solar-Energy} & \multicolumn{4}{c}{Electricity} &
		\multicolumn{4}{c}{Exchange Rate}& \multicolumn{4}{c}{Wind}\\
		\cline{3-18}
		& &3 &6 &12 & 24 & 3 &6 &12 & 24 &3 &6 &12 & 24 &3 &6 &12 & 24 \\
		\hline
		\hline
		\multirow{2}{*}{AR} & RSE&0.2435 &0.3790 &0.5911 &0.8699  &0.0995 &0.1035 &0.1050 &0.1054 &0.0228 &0.0279 &0.0353 &0.0445 &0.7161 &0.7572 &0.8076 &0.9371  \\
		& CORR&0.9710 &0.9263 &0.8107 &0.5314 &0.8845 &0.8632 &0.8591 &0.8595 &0.9734 &0.9656 &0.9526 &0.9357 &0.6459 &0.6046 &0.5560 &0.4633  \\
		\hline
		\multirow{2}{*}{GP} & RSE&0.2259 &0.3286 &0.5200 &0.7973 &0.1500 &0.1907 &0.1621 &0.1273 &0.0239 &0.0272 &0.0394 &0.0580 &0.6689 &0.6761 &0.6772 &0.6819   \\
		& CORR&0.9751 &0.9448 &0.8518 &0.5971 &0.8670 &0.8334 &0.8394 &0.8818 &0.8713 &0.8193 &0.8484 &0.8278 &0.6964 &0.6877 &0.6846 &0.6781  \\
		\hline
		\multirow{2}{*}{VARMLP} & RSE&0.1922 &0.2679 &0.4244 &0.6841 &0.1393 &0.1620 &0.1557 &0.1274 &0.0265 &0.0394 &0.0407 &0.0578 &0.7356 &0.7769 &0.8071 &0.8334   \\
		& CORR&0.9829 &0.9655 &0.9058 &0.7149 &0.8708 &0.8389 &0.8192 &0.8679 &0.8609 &0.8725 &0.8280 &0.7675 &0.6415 &0.5973 &0.5724 &0.5470  \\
		\hline
		\multirow{2}{*}{RNN-GRU} & RSE&0.1932 &0.2628 &0.4163 &0.4852 &0.1102 &0.1144 &0.1183 &0.1295 &0.0192 &0.0264 &0.0408 &0.0626 &0.6131 &0.6479 &0.6573 &0.6381 \\
		& CORR&0.9823 &0.9675 &0.9150 &0.8823 &0.8597 &0.8623 &0.8472 &0.8651 &0.9786 &0.9712 &0.9531 &0.9223 &0.7403 &0.7089 &0.6956 &0.7173  \\
		\hline
		\multirow{2}{*}{LSTNet} & RSE&0.1843 &0.2559 &0.3254 &0.4643 &0.0864 &0.0931 &0.1007 &0.1007 &0.0226 &0.0280 &0.0356 &0.0449 &\textbf{0.6079} &0.6262 &0.6279 &\textbf{0.6257}   \\
		& CORR&0.9843 &0.9690 &0.9467 &0.8870 &0.9283 &0.9135 &0.9077 &0.9119 &0.9735 &0.9658 &0.9511 &0.9354 &\textbf{0.7436} &0.7275 &0.7249 &\textbf{0.7284}  \\
		\hline
		\multirow{2}{*}{TPA-LSTM} & RSE&0.1803 &0.2347 &0.3234 &0.4389 &0.0823 &0.0916 &0.0964 &0.1006 &\textbf{0.0174} &\textbf{0.0241} &\textbf{0.0341} &\textbf{0.0444} &0.6093 &0.6292 &0.6290 &0.6335   \\
		& CORR&0.9850 &0.9742 &0.9487 &0.9081 &0.9439 &0.9337 &0.9250 &0.9133 &0.9790 &0.9709 &0.9564 &0.9381 &0.7433 &0.7240 &0.7235 &0.7202  \\
		\hline
		\multirow{2}{*}{MTGNN} & RSE&0.1778 &0.2348 &0.3109 &0.4270 &0.0745 &0.0878 &0.0916 &\textbf{0.0953} &0.0194 &0.0259 &0.0349 &0.0456 &0.6204 &0.6346 &0.6363 &0.6426   \\
		& CORR&0.9852 &0.9726 &0.9509 &0.9031 &0.9474 &0.9316 &0.9278 &0.9234 &0.9786 &0.9708 &0.9551 &0.9372 &0.7337 &0.7209 &0.7164 &0.7134  \\
		\hline
		\multirow{2}{*}{StemGNN} & RSE &0.1839 &0.2612 &0.3564 &0.4768 &0.0799 &0.0909 &0.0989 &0.1019 &0.0506 &0.0674 &0.0676 &0.0685 &0.6197 &0.6358 &0.6243 &0.6379   \\
		& CORR &0.9841 &0.9679 &0.9395 &0.8740 &0.9490 &\textbf{0.9397} &\textbf{0.9342} &0.9209 &0.8871 &0.8703 &0.8499 &0.8738 &0.7282 &0.7202 &0.7228 &0.7130  \\
		\hline
		\hline
		\multirow{2}{*}{\textbf{ESG}} & RSE&\textbf{0.1708} &\textbf{0.2278} &\textbf{0.3073} &\textbf{0.4101} &\textbf{0.0718} &\textbf{0.0844} &\textbf{0.0898} &0.0962 &0.0181 &0.0246 &0.0345 &0.0468 &0.6118 &\textbf{0.6250} &\textbf{0.6272} &0.6298\\
		& CORR&\textbf{0.9865} &\textbf{0.9743} &\textbf{0.9519} &\textbf{0.9100} &\textbf{0.9494} &0.9372 &0.9321 &\textbf{0.9279} &\textbf{0.9792} &\textbf{0.9717} &\textbf{0.9564} &\textbf{0.9392} &0.7417 &\textbf{0.7281} &\textbf{0.7258} &0.7225 \\
		\hline
	\end{tabular}
	}
	\label{tab:singlemain}
\end{table*}

\begin{table*}[tbh!]           
	\centering
	\caption{Comparison with baselines on multi-step forecasting.}
	\resizebox{\textwidth}{!}{
	\begin{tabular}{cc||ccc|ccc|ccc|ccc}
		\hline
		\multirow{2}{*}{Dataset} &
		\multirow{2}{*}{Method} & \multicolumn{3}{c}{Horizon 3} & \multicolumn{3}{c}{Horizon 6} &
		\multicolumn{3}{c}{Horizon 12} &
		\multicolumn{3}{c}{All}\\
		\cline{3-14}
		& & RMSE & MAE & CORR & RMSE & MAE & CORR & RMSE & MAE & CORR & RMSE & MAE & CORR\\
		\hline
		\hline
	    \multirow{8}{*}{NYC-Bike}
		&XGBoost  &3.7048  &2.2167 &0.5232  &4.1747 &2.5511 &0.3614 &4.3925 &2.7091 &0.2894 &4.0494  &2.4689 &0.4107    \\
		&DCRNN  &3.0172  &1.7917 &0.6967  &3.2369 &1.9078 &0.6609 &3.5100 &2.0325 &0.6196 &3.2274  &1.8973 &0.6601    \\
		&STGCN  &2.6256  &1.6456 &0.7539  &3.8368 &2.2827 &0.6282 &4.3713 &2.6052 &0.4521 &3.7829  &2.2076 &0.5933    \\
		&STG2Seq &3.4669  &2.0409 &0.5999  &3.9145 &2.2630 &0.5079 &4.2373 &2.5163 &0.4443 &3.7843  &2.2055 &0.5413    \\
		&STSGCN  &2.7328  &1.6973 &0.7386  &2.8861 &1.7416 &0.7179 &3.0548 &1.8224 &0.6903 &2.8846  &1.7538 &0.7126    \\
		&MTGNN  &2.5962  &1.5668 &0.7626  &2.7588 &1.6525 &0.7447 &3.3068 &1.7892 &0.6931 &2.7791  &1.6595 &0.7353   \\
		&CCRNN  &2.6538  &1.6565 &0.7534 &2.7561 &1.7061 &0.7411 &2.9436 &1.8040 &0.7029 &2.7674  &1.7133 &0.7333   \\
		&GTS  &2.7628  &1.7159 &0.7248  &2.9287 &1.7769 &0.7007 &3.1649 &1.8905 &0.6622 &2.9258  &1.7798 &0.6985   \\
		\hline
		&\textbf{ESG} &\textbf{2.5529} &\textbf{1.5483} &\textbf{0.7638} &\textbf{2.6484} &\textbf{1.6026} &\textbf{0.7511} &\textbf{2.8778} &\textbf{1.7173} &\textbf{0.7152} &\textbf{2.6727} &\textbf{1.6129} &\textbf{0.7449} \\
		\hline
		\hline
	    \multirow{8}{*}{NYC-Taxi}
		&XGBoost  &15.0372  &8.4121 &0.6862  &21.3395 &11.8491 &0.4433 &26.7073 &15.7165 &0.0452 &21.1994  &11.6806 &0.4416    \\
		&DCRNN  &12.3223  &7.0655 &0.7591  &15.1599 &8.6639 &0.6634 &17.8194 &10.5095 &0.5395 &14.8318  &8.4835 &0.6671    \\
		&STGCN  &11.2175  &6.1441  &0.8090  &14.0360 &7.6797 &0.7470 &18.7168 &10.2211 &0.5922 &14.6473  &7.8435 &0.7257    \\
		&STG2Seq &14.0756  &7.7274 &0.7258  &19.1757 &10.5066 &0.5429 &24.5691 &14.3603 &0.2855 &19.2077  &10.4925 &0.5389    \\
		&STSGCN  &10.5381  &5.6448 &0.8370  &10.8444 &5.7634 &0.8302 &11.9443 &6.3185 &0.7988 &10.9692 &5.8299 &0.8242    \\
		&MTGNN  &10.3394  &5.6775 &0.8374  &10.7534 &5.8168 &0.8312 &12.5164 &6.5285 &0.7972 &10.9472  &5.9192 &0.8249     \\
		&CCRNN  &9.3033  &5.4586 & 0.8529  &9.7794 & 5.6362 &0.8438 &10.9585 &6.1416 & 0.8186 &9.8744  &5.6636 & 0.8416   \\
		&GTS    &10.7796  &6.2337 &0.7974  &13.0215 &7.3251 &0.7299 &14.9906 &8.5328 &0.6524 &12.7511  &7.2095 &0.7348   \\
		\hline
		&\textbf{ESG} &\textbf{8.5745} &\textbf{4.8750} &\textbf{0.8656} &\textbf{9.0125} &\textbf{5.0500} &\textbf{0.8592} &\textbf{9.7857} &\textbf{5.4019} &\textbf{0.8450} &\textbf{8.9759} &\textbf{5.0344} &\textbf{0.8592} \\
		\hline
	\end{tabular}
	}
	\label{tab:multimain}
\end{table*}

\subsection{Baselines}
We utilize two groups of baselines for single-step and multi-step forecasting respectively. Both empirical statistic methods and popular deep learning models are selected carefully. 
The detailed information of baselines is shown in Appendix \ref{sec:baseline}.

\textbf{Single-step forecasting.} We select 8 time-series forecasting methods. Two empirical statistical methods are Auto-Regressive (AR), Gaussian Process (GP) \cite{roberts2013gaussian}. Six deep learning models contain VARMLP \cite{zhang2003time}, RNN-GRU, LSTNet \cite{lai2018modeling}, TPA-LSTM \cite{shih2019temporal}, MTGNN \cite{wu2020connecting}, and StemGNN \cite{cao2021spectral}.

\textbf{Multi-step forecasting.} 8 popular baselines are selected including XGBoost \cite{chen2016xgboost}, DCRNN \cite{li2017diffusion}, STGCN \cite{yu2018spatio}, STG2Seq \cite{bai2019stg2seq}, STSGCN \cite{song2020spatial}, MTGNN \cite{wu2020connecting}, CCRNN \cite{ye2021coupled}, and GTS \cite{shang2021discrete}.

\subsection{Main Results}

Table \ref{tab:singlemain} and Table \ref{tab:multimain} summarize the single-step and multi-step forecasting evaluation results. In summary, ESG achieves state-of-the-art performance in both two tasks. The best results are highlighted in bold font. 

\subsubsection{Single-step Forecasting}
In this task, we compare ESG with other multivariate time series models. Table \ref{tab:singlemain} shows the detailed experimental results on the single-step forecasting.
Excellent performance of our model is achieved on Solar-Energy, Electricity. Especially on Solar-Energy dataset, ESG achieves 3.94$\%$, 3.96$\%$ improvements compared with the state-of-the-art methods on the horizons of 3, 24 in terms of RSE.
On this dataset, the excellent results are also achieved by MTGNN which utilizes the self-learned adjacency matrix to describe the correlations between time series. The impact of weather on the power generation is shared between the plants in the same areas, which offers the static graph a strong local relationship. However, ESG still makes a significant improvement. This is because the evolving correlations between multivariate time series are well captured in Solar-Energy dataset.
On the Exchange-rate and Wind datasets, the results of ESG are not as good as results on the first two datasets, but ESG still achieves optimal performance on more than half of the metrics. This is possibly due to the smaller graph size and fewer training examples of Exchange-rate and Wind datasets.

\subsubsection{Multi-step Forecasting}
In this task, several spatio-temporal methods are chosen in the traffic prediction domain. Table \ref{tab:multimain} shows the detailed experimental results on the multi-step forecasting. The performance on time steps 3, 6, 12, and the average of all horizons is compared. 
In general, our ESG achieves state-of-the-art results regarding all the metrics for all horizons on both two datasets. In particular, ESG lowers down RMSE by 7.8$\%$, 7.8$\%$, 10.7$\%$ and MAE by 10.7$\%$, 10.4$\%$, 12.0$\%$ over the horizons of 3, 6, 12 on the NYC-taxi data. The improvement of ESG increases with the growth of the forecasting horizon, indicating our model ability in long-term forecasting, which is inherently more uncertain and difficult than short-term forecasting.
We can observe further phenomena from the table. XGBoost which doesn't explore the correlations between time series obtains the worst results. DCRNN, STGCN, and STG2Seq are all graph convolution network-based forecasting models, which outperform XGBoost. However, the graph structures they utilize in GCN are fixed and hand-crafted, which is not flexible and representative enough to describe the correlations. MTGNN, CCRNN, and GTS adopt an adaptive static graph structure learning manner, which contributes to their significant improvement over previous methods. 
However, being lack of extracting evolving correlations restricts their further improvements.

\subsection{Ablation Study}
To validate the effectiveness of the key components, we conduct a ablation study on NYC-Bike. We name variants of ESG as follows: 
\begin{itemize}
    \item \textbf{Static Graph Only}: Removing the evolving graph structure, utilizing the static graph constructed by static node representation only.
    \item \textbf{w/o Scale-Specific}: Generating a series of evolving graph structures from the raw input data only, which is later used in each mix-hop propagation layer regardless of changes in correlations over different time scales.
    \item \textbf{Same Pattern of Evolution}: Sharing parameters of evolving graph structure learners at different scales.
\end{itemize}

We repeat each experiment 10 times and report the average value and the standard deviation in Table \ref{tab:ablation}. More details are shown in Appendix \ref{sec:extraexp}.  We could make a conclusion that all components contribute to the final state-of-the-art results to a certain extent. 
Firstly, removing the evolving graph structure learner still obtains a competitive result, which indicates that the robust and informative self-learned graph structure could help a lot in multivariate time series forecasting. However, only using the static graph, without considering the evolution of the graph structure, brings a large standard deviation to the results.
Secondly, using dynamic graphs regardless of the difference of scales reaches the worst results. This is because the graph structure varies at different time scales. In addition, the dynamic graphs generated in this common manner cannot correspond to the information processed by the temporal convolution module in the time dimension. The fact verifies the necessity and effectiveness of using the scale-specific evolving graph structure learner.
The comparison between the third variant and ESG proves the correctness of the hypothesis that the evolution patterns of graph structure follows vary at different time scales.

\begin{figure}[tb]
	\centering
	\includegraphics[width=\columnwidth]{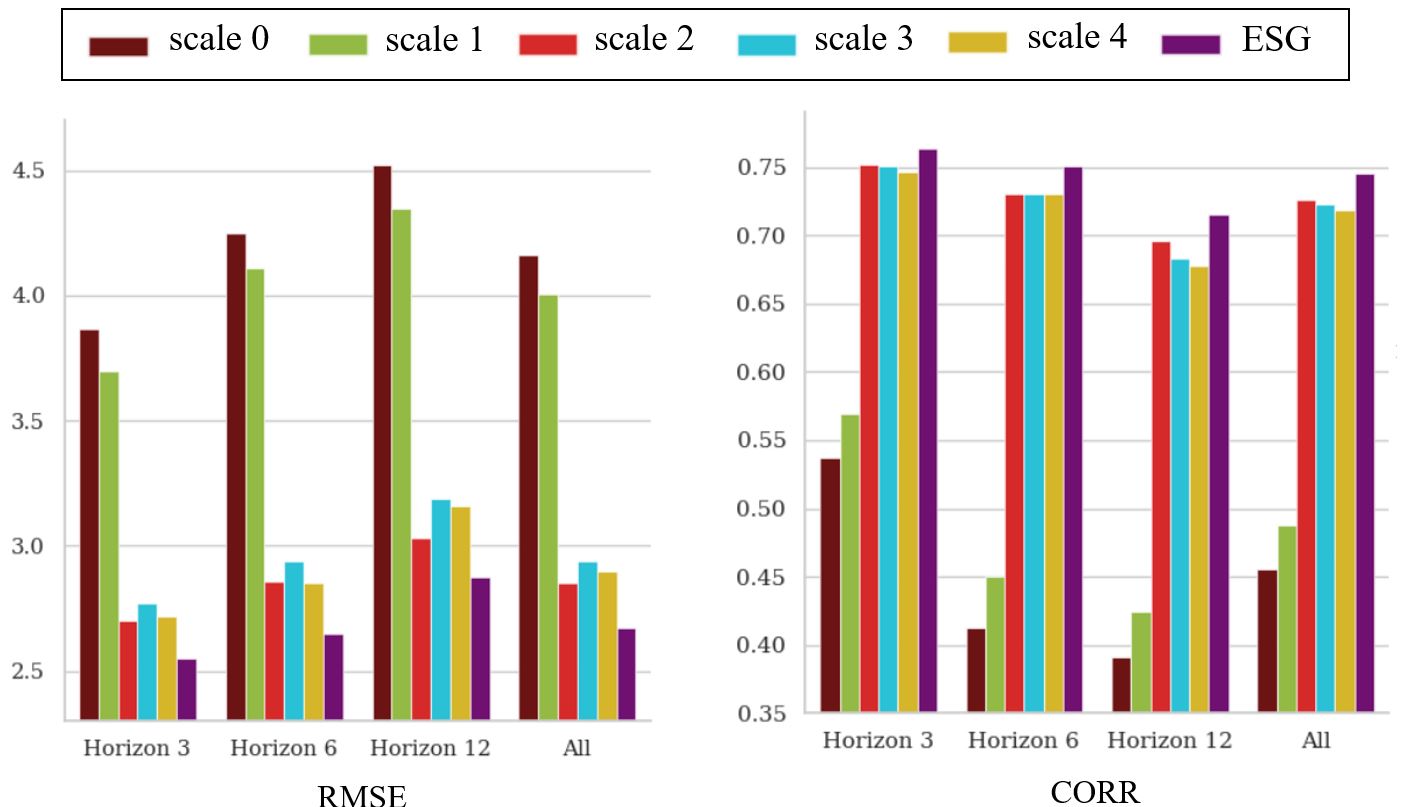}
	\caption{Utilizing the information at different scales.}
	\label{fig:diffscale}
\end{figure}

To further investigate the importance of multi-scale information fusion,
we evaluate the performance of ESG variants which only employ one scale information, and the results are visualized as the barplot shown in Figure \ref{fig:diffscale}. In the experiments, scale 0 denotes using raw input. Scale 1, scale 2, and scale 3 use the output of the first, second, and third temporal convolution module respectively, which represent different time scales from the short to long term. Scale 4 uses the output of the last graph convolution module. ESG outperforms the methods using only one scale information by a large margin, which indicates the superiority of fusing the multi-scale representations to make the final prediction. Interestingly, scale 2 achieves the second-best results. It suggests that the importance of different scales varies. Furthermore, the contribution to the forecasting doesn't weigh more as the scale level increases.

\begin{table}[tb]           
	\centering
	\caption{Ablation Study.}
	\resizebox{\columnwidth}{!}{
	\begin{tabular}{c||c|c|c}
	\hline
		Method & RMSE & MAE & CORR\\
		\hline
		\hline
		Static Graph Only  &2.7439$\pm$0.0438 	&1.6302$\pm$0.0176 	&0.7388$\pm$0.0050 \\	
		w/o Scale-Specific  &2.8102$\pm$0.0433 &1.6663$\pm$0.0150 &0.7259$\pm$0.0047 \\	
		Same Pattern of Evolution  &2.7274$\pm$0.0177 &1.6296$\pm$0.0036 &0.7402$\pm$0.0024\\ 
		\hline
		\textbf{ESG}  &\textbf{2.6727$\pm$0.0117} &\textbf{1.6129$\pm$0.0086} &\textbf{0.7449$\pm$0.0051} \\
		\hline
	\end{tabular}
	}
	\label{tab:ablation}
\end{table}

\subsection{Study of Evolving Graph Structure}

\begin{figure*}[tbh!]
	\centering
	\includegraphics[width=2\columnwidth]{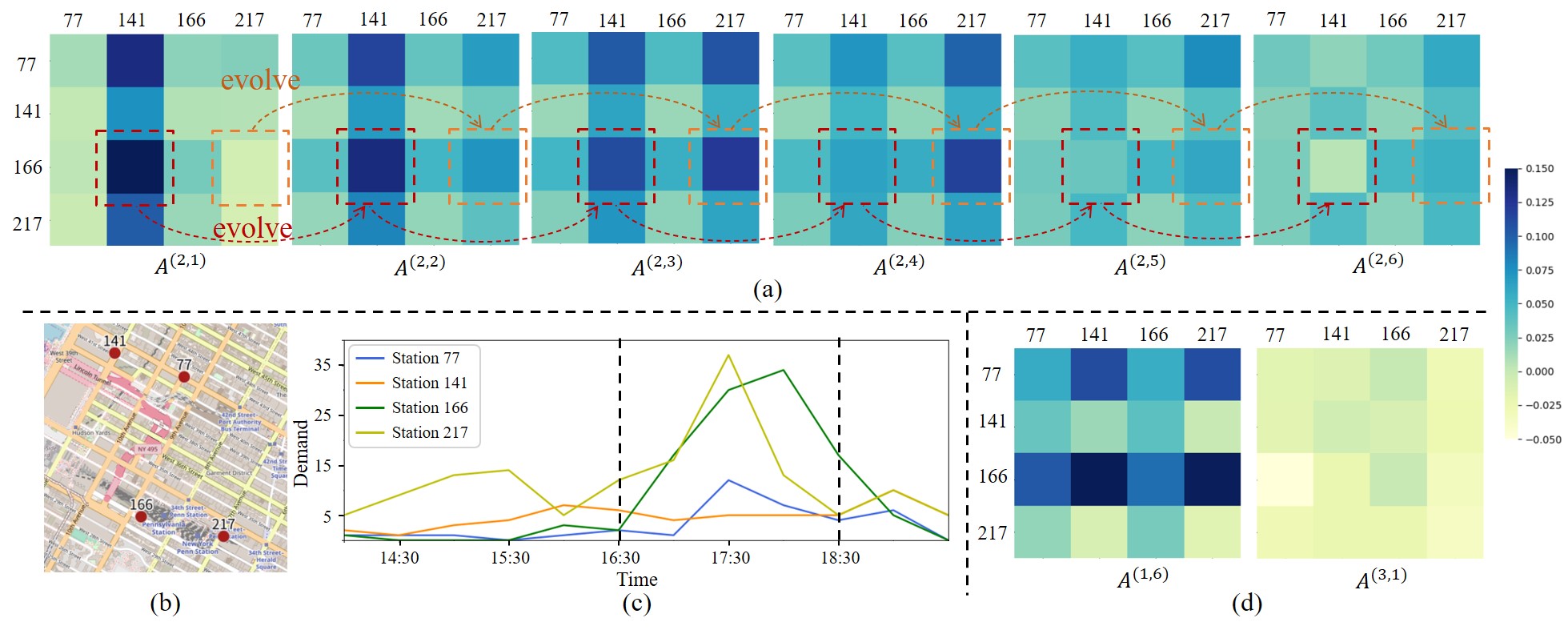}
	\caption{(a) A series of adjacency matrices in scale 2 on the NYC-Bike dataset, which reveals a strong evolving pattern. (b) The location of node 77, 141, 166 and 217 on the map. (c) The raw time series curves on 12 time steps, which corresponds to the adjacency matrices shown in (a) and (d). (d) Several adjacency matrices on scale 1 and 3.}
	\label{fig:study1}
\end{figure*}

To further verify the effectiveness of evolving graph structure learner, we select four stations with number 77, 141, 166, and 217 on April $8th$, 2016 to conduct a practical case study.
As it is shown in Figure \ref{fig:study1} (a), $\bm{A}^{(2,1)}, \bm{A}^{(2,2)},...,$ and $\bm{A}^{(2,6)}$ are the adjacency matrices at scale 2, which are visualized as heat map. The bluer grid indicates a bigger weight.  Figure \ref{fig:study1} (b) and Figure \ref{fig:study1} (c) display the practical locations and the raw time series curves.
Considering that numerous variables contain the genuine unidirectional relationship behind, we have not constrained the symmetry when generating the adjacency matrices.
Taking station 166 as an example, several interesting phenomena are observed. 1) Before 16:30 in Figure \ref{fig:study1} (c), we could observe that station 166, the green line, and station 141, the orange line, have a strong correlation with each other, and they go up and down together. However, the situation changes after 16:30, station 141 remains stable but station 166 fluctuates dramatically. The fact that the correlations evolve from high to low is well captured by the adjacency matrices. As it is shown in Figure \ref{fig:study1} (a), the value at row 166 and column 141 gets smaller and smaller over time. 2) As we could observe in Figure \ref{fig:study1} (a), the value along the orange dashed line goes deep in the beginning, and then it turns light after $\bm{A}^{(2,4)}$, which indicates the correlation between station 166 and 217 rises in the beginning and falls in the end. This information is also consistent with the fact shown in Figure \ref{fig:study1} (c). The yellow line has a different tendency from the green line in the first. Then they go up together from 16:30 to 18:30 and end up separating. The two phenomena above provide strong support for demonstrating the effectiveness of evolving graph structure learner. 
Additionally, though station 217 is closest to 166 in physical distance, the correlation between them is small before 16:30, which reflects the limitations of hand-crafted graph structure for its inflexibility and inaccuracy.

We also verify the graph structure which captures the the correlations at different observation scales. The total layer $L$ for the NYC-Bike dataset is 3. All adjacency matrices in scale 2 are shown in Figure \ref{fig:study1} (a), and we random select one in scale 1. Due to the sequence reducing led by the dilated convolution, the scale 3 only contain a single adjacency matrix which is also chosen. \ref{fig:study1} (d) display the two adjacency matrices, $\bm{A}^{(1,6)}$ and $\bm{A}^{(3,1)}$, in scale 1 and 3. We could observe that the values in the adjacency matrix at the scale 1 tend to be highly polarized, which indicates the short-term dependency of the stations is more likely to differ from others. However, at the last scale, the more average values in the adjacency matrix $\bm{A}^{(3,1)}$ clarify that the 4 time series possess the same pattern from the long-term view. 
The two case studies above offer us a strong support to verify that the evolving and multi-scale correlations among multivariate time series are well captured by ESG.

%% file: 5-Relatedwork.tex
\section{Related Work}
\label{sec:rel}
In this section, we review the related work from two perspectives, time series forecasting, and graph neural network.
\subsection{Time Series Forecasting}
Time series forecasting has attracted numerous researchers to dig into it. According to the number of variables that are taken as the observation, this enduring topic usually falls into two mainstreams, univariate time series, and multivariate time series. The former feeds the variables one by one to a single model under the assumption that all variables of the time series share the same temporal pattern. In the statistical domain, autoregressive integrated moving average (ARIMA) \cite{box2015time} is one of the most popular methods for its flexibility and excellent mathematical properties.
Recent years have witnessed the rapid growth of deep learning, and a number of researchers analyze the time series respectively under this framework. FC-LSTM \cite{sutskever2014sequence} combines the fully connected layer with LSTM to make prediction.
N-BEATS \cite{oreshkin2019n} utilizes deeply stacked fully-connected layers cooperated with the residual links, and the simple architecture pays back a great number of desired properties such as being interpretable and fast to train.

On the other side, multivariate time series forecasting takes the whole variables as an entity, and researchers devote themselves to exploring the correlation among the different time series. LSTNet \cite{lai2018modeling} and TPA-LSTM \cite{shih2019temporal} combined the convolution neural network and recurrent neural network to handle this problem in the deep learning framework. 
In the spatio-temporal prediction, the subfield of multivariate time series analysis, the similar model architecture is also preferred with treating the transportation demand of the whole city as several grids \cite{zhang2017deep,ye2019co}. 
DeepGLO \cite{sen2019think} decomposes the time series as the combination of the basis via matrix factorization (MF). TLAE \cite{nguyen2021temporal} followed this line and proposed a temporal auto-encoder. However, it is difficult for the methods above to capture the pair-wise correlation between variables precisely. To solve this problem, MTGNN \cite{wu2020connecting} generalized graph neural network to multivariate time series forecasting via a self-learned adjacency matrix. 

\subsection{Graph Neural Network}
\label{sec:rewo,gnn}
Recent years have witnessed the rapid development of graph neural network. \citet{bruna2013spectral} first extended the convolution network to graph-based data. The following work could be roughly divided into two categories \cite{wu2020survey}, spectral-based methods and spatial-based methods. The former one takes the graph convolution as the low-pass filter to remove high frequency information from the signals \cite{chebnet2016}.
Along this line, \citet{xu2018powerful} proved the expressiveness of graph neural networks is equal to Weisfeiler-Lehman (WL) graph isomorphism test. The spatial-based methods follow the message passing rule and aggregate the information from neighbors \cite{kipf2016gcn,hamilton2017inductive}.

With the rapid development of the GNN's theory, the flexible architecture and strong generalization ability make it widely applied in numerous domains. For example, in spatio-temporal data prediction, especially in traffic forecasting problem, the natural topological structure of the road network makes GNN achieve the remarkable performance \cite{li2017diffusion,yu2018spatio,guo2019attention,ye2021coupled}. 
\citet{yu2018spatio} first combined the GNN with the 1D convolution neural network and constructed the adjacency matrix by the distance between nodes.
\citet{guo2019attention} aggregated the recent, daily, weekly representations to give the final forecasting.
However, the evolving and multi-scale correlations among multivariate time series are seldom captured.

%% file: 6-Conclusion.tex
\section{Conclusion}
\label{sec:con}
In this paper, we proposed a novel multivariate time series forecasting model named ESG. In particular, an evolving graph structure learner is proposed to construct a series of adjacency matrices that not only receive the information from current input but also maintain the hidden states from the historical graph structure. Based on this, a hierarchical architecture is proposed to capture the multi-scale inter-and intra-time-series correlations simultaneously.
Finally, a unified forecasting framework integrates the components above to give the final prediction. 
The extensive experiments conducted on real-world datasets demonstrate the superiority of the proposed methods over the baselines. This research provides a new perspective to the correlation modeling on current multivariate time series forecasting. In the future, the evolving graph structure will be explored in more scenarios.

%% file: Acknowledgements.tex
\section*{Acknowledgements}
This work was supported by the National Natural Science Foundation of China (71901011, U1811463, 51991391, 51991395, U21A20516).

%% file: 7-Appendix.tex
\clearpage
\appendix

\section{Appendix}
More details are displayed in this section.
\subsection{Dataset and Metrics}
\label{sec:data}
We conduct detailed experiments on two groups of datasets, 4 for the single-step, and 2 for the multi-step task. Now more details of each dataset and the evaluation metrics are shown in the following.
\subsubsection{Single-Step Forecasting.}
4 datasets are conducted for this task, covering the energy, and exchange rates of countries.
\begin{itemize}
    \item \textbf{Solar-Energy}: This dataset \cite{lai2018modeling} from National Renewable Energy Laboratory contains the solar power production records of 137PV plants in Alabama State in 2007.
    \item \textbf{Electricity}: This dataset \cite{lai2018modeling} which is published by National Renewable Energy Laboratory contains the hourly electricity consumption of 321 clients from 2012 to 2014.
    \item \textbf{Exchange Rate}: This dataset \cite{lai2018modeling} contains daily exchange rates of eight countries from 1990 to 2016. The countries are Australia, British, Canada, Switzerland, China, Japan, New Zealand, and Singapore.
    \item \textbf{Wind}: This dataset \cite{wu2020adversarial} contains the hourly energy potential estimates of an area from 1986 to 2015.
\end{itemize}
The length of the look-back window $P$ is 168. For each future horizon ($Q=3,6,12,24$), the model is trained independently. The feature dimension $C$ is 1. The performance of models is evaluated by the Root Relative Squared Error (RSE, defined in Equation \eqref{rse}) and Empirical Correlation Coefficient (CORR, defined in Equation \eqref{corr}). The $\rho$ indicates the total number of samples, and $N$ is the number of nodes. $\bm{Y}$ and$\hat{\bm{Y}}$ indicate the ground truth and forecasting value. $\bar{\bm{Y}}$ and $\bar{\hat{\bm{Y}}}_n$ represent the mean values.

\subsubsection{Multi-Step Forecasting.}
Two real-world traffic datasets published by New York OpenData are chosen for this task.
\begin{itemize}
    \item \textbf{NYC-Bike}: This dataset \cite{ye2021coupled} collects the sharing bike demand of the residents' daily usage at 250 bike stations in New York from April $1st$, 2016 to June $30th$, 2016.
    \item \textbf{NYC-Taxi}: This dataset \cite{ye2021coupled} contains the taxi demand data in New York from April $1st$, 2016 to June $30th$, 2016.
\end{itemize}
The look-back window is 12 (6 hours), and we predict the future values for the next 12 time steps (6 hours). The feature dimension $C$ is 2, i.e., the demand of pick-up and drop-off. The performance of models is evaluated by Root Mean Squared Error (RMSE, defined in Equation \eqref{rmse}), Mean Absolute Error (MAE, defined in Equation \eqref{mae}) and Empirical Correlation Coefficient (CORR, defined in Equation \eqref{corr}).

\begin{equation}
\label{rmse}
RMSE  = \sqrt{\sum^{\rho}_{t^{\prime}=0}(\bm{Y}^{(t^{\prime})}-\hat{\bm{Y}}^{(t^{\prime})})^2}.
\end{equation}
\begin{equation}
\label{mae}
MAE  = \sum^{\rho}_{t^{\prime}=0}|\bm{Y}^{(t^{\prime})}-\hat{\bm{Y}}^{(t^{\prime})}|.
\end{equation}
\begin{equation}
\label{rse}
RSE  = \frac{\sqrt{\sum^{\rho}_{t^{\prime}=0}(\bm{Y}^{(t^{\prime})}-\hat{\bm{Y}}^{(t^{\prime})})^2}}{\sqrt{\sum^{\rho}_{t^{\prime}=0}(\bm{Y}^{(t^{\prime})}-\bar{\bm{Y}})^2}}.
\end{equation}
\begin{equation}
\label{corr}
CORR = \frac{1}{N}\sum^{N}_{n=1} 
	\frac{\sum_{t^{\prime}=0}^{\rho}(\hat{\bm{Y}}_{n}^{(t^{\prime})} - \bar{\hat{\bm{Y}}}_n)(\bm{Y}_{n}^{(t^{\prime})} - \bar{\bm{Y}}_n)}
	{\sqrt{\sum_{t^{\prime}=0}^{\rho}(\hat{\bm{Y}}_{n}^{(t^{\prime})} - \bar{\hat{\bm{Y}}}_n)^2(\bm{Y}_{n}^{(t^{\prime})} - \bar{\bm{Y}}_n)^2}}.
\end{equation}

\subsection{Baselines}
\label{sec:baseline}
Two groups of baselines are chosen for single-step and multi-step forecasting respectively.
All key hyper-parameters are well-tuned to ensure their performance.
Each experiment is run 10 times and the average value is presented.
\subsubsection{Single-step forecasting.}
For the single-step forecasting task, we select 8 baselines covering classical statistic methods and recent deep neural networks.
\begin{itemize}
    \item \textbf{AR}: Auto-Regression model capture the linear correlations among time series.
    \item \textbf{GP}: Gaussian Process \cite{roberts2013gaussian} is employed for time series modeling.
    \item \textbf{VARMLP}: VARMLP \cite{zhang2003time} is a hybrid methodology that combines both VAR and MLP models.
    \item \textbf{RNN-GRU}: The gated recurrent unit is employed for time series modeling.
    \item \textbf{LSTNet}: LSTNet \cite{lai2018modeling} combines the CNN with RNN for multivariate time series forecasting.
    \item \textbf{TPA-LSTM}: TPA-LSTM \cite{shih2019temporal} utilizes attention mechanism and RNN.
    \item \textbf{MTGNN}: MTGNN \cite{wu2020connecting} employs an adaptive adjacency matrix to describe the correlation among time series.
    \item \textbf{StemGNN}: StemGNN \cite{cao2021spectral}  employs Fourier Transform to discover the hidden patterns of time series. 
\end{itemize}

\subsubsection{Multi-step forecasting.}
We choose 8 competitive and representative baselines for the multi-step forecasting task.
\begin{itemize}
    \item \textbf{XGBoost}: A powerful machine learning method based on gradient boosting tree \cite{chen2016xgboost}.
    \item \textbf{DCRNN}: Diffusion convolutioal recurrent neural network \cite{li2017diffusion} first utilizes a hand-craft adjacency matrix to describe the correlations among time series.
    \item \textbf{STGCN}: Sptaio-temporal graph convolutional network \cite{yu2018spatio} integrates the graph convolution with 1D convolution neural network.
    \item \textbf{STG2Seq}: Spatio-temporal graph to sequence model \cite{bai2019stg2seq} captures the long-term and short-term temporal dependency respectively.
    \item \textbf{STSGCN}: Spatial-temporal synchronous graph convolutional network \cite{song2020spatial} constructs a 3D graph convolution kernel.
    \item \textbf{MTGNN}: This method \cite{wu2020connecting} capture the correlations among variables via a self-learned adjacency matrix.
    \item \textbf{CCRNN}: Coupled layer-wise convolutional recurrent neural network \cite{ye2021coupled} explores the hierarchical graph convolution for multivariate time series.
    \item \textbf{GTS}: Graph for Time Series \cite{shang2021discrete} combines a discrete graph structure learner with recurrent neural network.

\end{itemize}

 \begin{table*}[tbh!]           
	\centering
	\caption{Ablation Study.}
	\resizebox{\textwidth}{!}{
	\begin{tabular}{c||ccc|ccc|ccc|ccc}
	\hline
		\multirow{2}{*}{Method} & \multicolumn{3}{c}{Horizon 3} & \multicolumn{3}{c}{Horizon 6} &
		\multicolumn{3}{c}{Horizon 12} &
		\multicolumn{3}{c}{All}\\
		\cline{2-13}
		& RMSE & MAE & CORR & RMSE & MAE & CORR & RMSE & MAE & CORR & RMSE & MAE & CORR\\
		\hline
		\hline

		Static Graph Only &2.6217 &1.5686 &0.7600 &2.7451 &1.6327	&0.7401 &2.9315 &1.7254 &0.7098 &2.7439	&1.6302	&0.7388 \\
	
		w/o Scale-specific  &2.6504 &1.5872 &0.7530   &2.7907 &1.6623 &0.7304 &3.0320 &1.7781 &0.6940 &2.8102 &1.6663 &0.7259 \\
		Same Pattern of Evolution &2.5911 &1.5593 &0.7612 &2.7179 &1.6263 &0.7443 &2.9224 &1.7297 &0.7114 &2.7274	&1.6296	&0.7402 \\

		\hline
		\textbf{ESG} &\textbf{2.5529} &\textbf{1.5483} &\textbf{0.7638} &\textbf{2.6484} &\textbf{1.6026} &\textbf{0.7511} &\textbf{2.8778} &\textbf{1.7173} &\textbf{0.7152} &\textbf{2.6727} &\textbf{1.6129} &\textbf{0.7449}  \\
		\hline
	\end{tabular}
	}
	\label{tab:totalablation}
\end{table*}

\subsection{Experimental Setup}
\label{sec:setup}
The models is implemented by the Pytorch. All the experiments are conducted on an Ubuntu machine equipped with two Intel(R) Xeon(R) CPU E5-2667 v4 @ 3.20GHz with 8 physical cores, and the GPU is NVIDIA TITAN Xp, armed with 12 GB of GDDR5X memory running at over 11 Gbps. 
We repeat the experiment 10 times and report the average value of evaluation metrics. The model is trained by the Adam optimizer with gradient clip 5. Learning rate is chosen from $\{$0.01, 0.005, 0.001, 0.0005, 0.0001$\}$ by grid search. Dropout with 0.3 is applied after each temporal
convolution module. Layernorm is applied after each graph convolution module. The depth of the mix-hop propagation layer $\Psi$ is set to 2. The retain ratio $\beta$ from the mix-hop propagation layer is set to 0.05. The dimension of static node representation $C_s$ is 40. Other hyper-parameters are reported according to different tasks.
\subsubsection{Single-step forecasting.}
We stack 5 multi-scale extractors with a sequence of time intervals $d^{(l)}$, 31, 31, 21, 14, 1.  In temporal convolution module, the dilation factor for each layer grows exponentially at a rate of 2 and four filter sizes are used, i.e., $k_1$=2, $k_2$=3, $k_3$=6, and $k_4$=7. The output channels of temporal convolution modules $C_{\bm{\xi}}$ and graph convolution modules $C_Z$ both are 16. The skip connection layers are $1\times T^{(l)}$ standard convolutions, where $T^{(l)}$ is the sequence length of the inputs to the $l$-th skip connection layer, which all have 32 output channels. In the output module, The first layer has 64 output channels and the second layer has 1 output channel. For the Solar-
Energy, the batch size is 16. For the Electricity and Exchange Rate, the batch size is 4. For the Wind, the batch size is 32. The dimension of evolving node representation $C_e$ is set to 20 for the Solar-Energy, Electricity, and Wind, while set to 16 for the Exchange Rate.

\subsubsection{Multi-step forecasting.}
 We stack 3 multi-scale extractors with a sequence of time intervals $d^{(l)}$, 1, 1, 1.  In temporal convolution module, the dilation exponential factor is 1 and two filter sizes are used, i.e., $k_1$=2, $k_2$=6. The output channels of temporal convolution modules $C_{\bm{\xi}}$ and graph convolution modules $C_Z$ both are 32. The skip connection layers are $1\times T^{(l)}$ standard convolutions, which all have 64 output channels. In the output module, The first layer has 128 output channels and the second layer has 12$\times$2 output channels. The batch size is 16 and the dimension of evolving node representation $C_e$ is set to 20.
 
\subsection{Extra Experiment Result}
\label{sec:extraexp}
We conduct a detailed ablation study on NYC-Bike dataset. As shown in Table \ref{tab:totalablation}, ESG achieves the best results on the average value, as well as on horizon 3, 6, 12.